\documentclass[conference]{IEEEtran}
\IEEEoverridecommandlockouts
\usepackage{cite}
\usepackage{amsmath,amssymb,amsfonts}
\usepackage{algorithmic}
\usepackage{graphicx}
\usepackage{textcomp}
\usepackage{xcolor}
\def\BibTeX{{\rm B\kern-.05em{\sc i\kern-.025em b}\kern-.08em
    T\kern-.1667em\lower.7ex\hbox{E}\kern-.125emX}}
\begin{document}

\title{Employing Drones in Agriculture: An Exploration of Various Drone Types and Key Advantages
}

\author{\IEEEauthorblockN{1\textsuperscript{st}  Eduardo Carvalho Nunes}
\IEEEauthorblockA{\textit{Department of Engineering} \\
\textit{University of Trás-os-Montes and Alto Douro}\\
5000-801, Vila Real, Portugal \\
ORCID: 0000-0002-5345-8854}

}

\maketitle

\begin{abstract}

This article explores the use of drones in agriculture and discusses the various types of drones employed for different agricultural applications. Drones, also known as unmanned aerial vehicles (UAVs), offer numerous advantages in farming practices. They provide real-time and high-resolution data collection, enabling farmers to make informed irrigation, fertilization, and pest management decisions. Drones assist in precision spraying and application of agricultural inputs, minimizing chemical wastage and optimizing resource utilization. They offer accessibility to inaccessible areas, reduce manual labor, and provide cost savings and increased operational efficiency. Drones also play a crucial role in mapping and surveying agricultural fields, aiding crop planning and resource allocation. However, challenges such as regulations and limited flight time need to be addressed. The advantages of using drones in agriculture include precision agriculture, cost and time savings, improved data collection and analysis, enhanced crop management, accessibility and flexibility, environmental sustainability, and increased safety for farmers. Overall, drones have the potential to revolutionize farming practices, leading to increased efficiency, productivity, and sustainability in agriculture.

\end{abstract}


\section{Introduction}

The use of drones in agriculture has gained significant attention in recent years due to their potential to revolutionize farming practices. Drones, also known as unmanned aerial vehicles (UAVs), offer a range of applications that can enhance efficiency, productivity, and sustainability in agriculture. 

One of the key advantages of using drones in agriculture is their ability to provide real-time and high-resolution data collection \cite{10.1002/net.21818}. Drones equipped with cameras, sensors, and imaging technologies can capture detailed imagery of crops, soil conditions, and field topography \cite{10.3390/rs9010088}. This data can be used for crop monitoring, assessment, and precision agriculture practices \cite{10.1109/access.2019.2949703}. By analyzing this data, farmers can make informed decisions regarding irrigation, fertilization, and pest management, leading to optimized resource utilization and improved crop yields \cite{10.3390/s20051487}. 

Drones also play a crucial role in precision spraying and application of agricultural inputs \cite{10.1109/access.2019.2949703}. With their ability to navigate through fields and deliver targeted treatments, drones can reduce chemical wastage, minimize environmental impact, and improve the efficiency of pesticide and fertilizer application \cite{10.1109/access.2019.2949703}. This targeted approach helps protect beneficial insects, reduce water pollution, and optimize resource utilization \cite{10.1109/access.2019.2949703}. 

Furthermore, drones offer accessibility to inaccessible or inaccessible areas by traditional means \cite{10.5937/ekonomika1804091s}. They can fly at low altitudes and capture data from different angles and perspectives, providing a comprehensive view of the field \cite{10.5937/ekonomika1804091s}. This enables farmers to monitor large farmland areas quickly and efficiently, reducing the time and labor required for manual inspections \cite{10.1002/net.21818}. Drones can cover large farmland areas in a fraction of the time it would take using traditional methods, leading to cost savings and increased operational efficiency \cite{10.1002/net.21818}. 

In addition to data collection and monitoring, drones can assist in mapping and surveying agricultural fields. They can create high-resolution maps and 3D models, providing valuable information for crop planning, land management, and resource allocation. Drones equipped with advanced sensors, such as LiDAR or hyperspectral cameras, can capture detailed data for precise analysis and decision-making \cite{10.3390/rs9010088}. This enables farmers to identify areas of nutrient deficiencies, optimize irrigation practices, and implement site-specific management strategies. The use of drones in agriculture is challenging. Regulations and licensing requirements for drone operation vary across countries and regions, and compliance with these regulations is essential to ensure safe and responsible drone use \cite{10.3390/land10020164}. Additionally, drones' limited flight time and battery capacity can pose challenges in large-scale farming operations \cite{10.3390/s20051487}. However, advancements in drone technology, such as improved battery life and payload capacity, are addressing these limitations and expanding the possibilities for drone applications in agriculture.

\section{Different Types of Drones used in Agriculture}
In agriculture, different types of drones are used for various applications. These drones offer unique capabilities and functionalities that cater to specific agricultural needs. Some of the commonly used types of drones in agriculture include: 

\begin{itemize}
    \item \textbf{Multi-Rotor Drones}: Multi-rotor drones (Figure \ref{fig:multi_example}), such as quadcopters and hexacopters, are popular in agriculture due to their maneuverability and stability \cite{10.1002/net.21818}. They are equipped with multiple rotors that allow them to hover in place, fly at low altitudes, and capture high-resolution imagery. Multi-rotor drones are suitable for tasks that require close and contained object capture, such as monitoring crop health, detecting pests and diseases, and applying targeted treatments \cite{10.3390/s20051487}.

    \begin{figure}[htbp]
    \centering
    \includegraphics[width=\columnwidth]{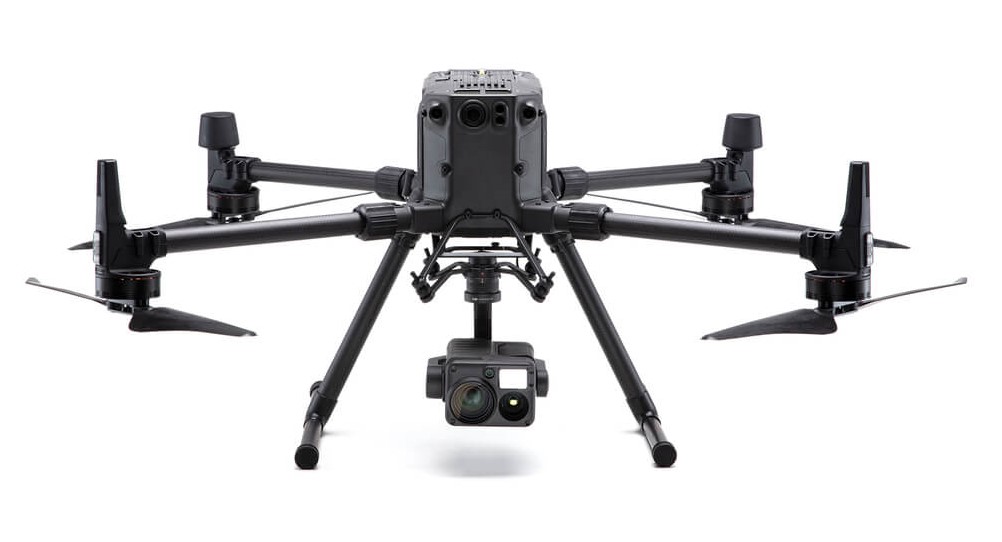}
    \caption{Example Multi-Rotor Drone - DJI Matrice 300 RTK}
    \label{fig:multi_example}
    \end{figure}

    \item \textbf{Fixed-Wing Drones}: Fixed-wing drones (Figure \ref{fig:fixed_example}) have a wing-like structure and are designed to fly like airplanes \cite{10.3390/drones6070160}. They are known for their long-flight endurance and ability to cover large areas. Fixed-wing drones are commonly used for mapping and surveying agricultural fields, as they can fly faster and cover more considerable distances. However, they require a runway for takeoff and landing, which can be a limitation in specific agricultural settings.

    \begin{figure}[htbp]
    \centering
    \includegraphics[width=\columnwidth]{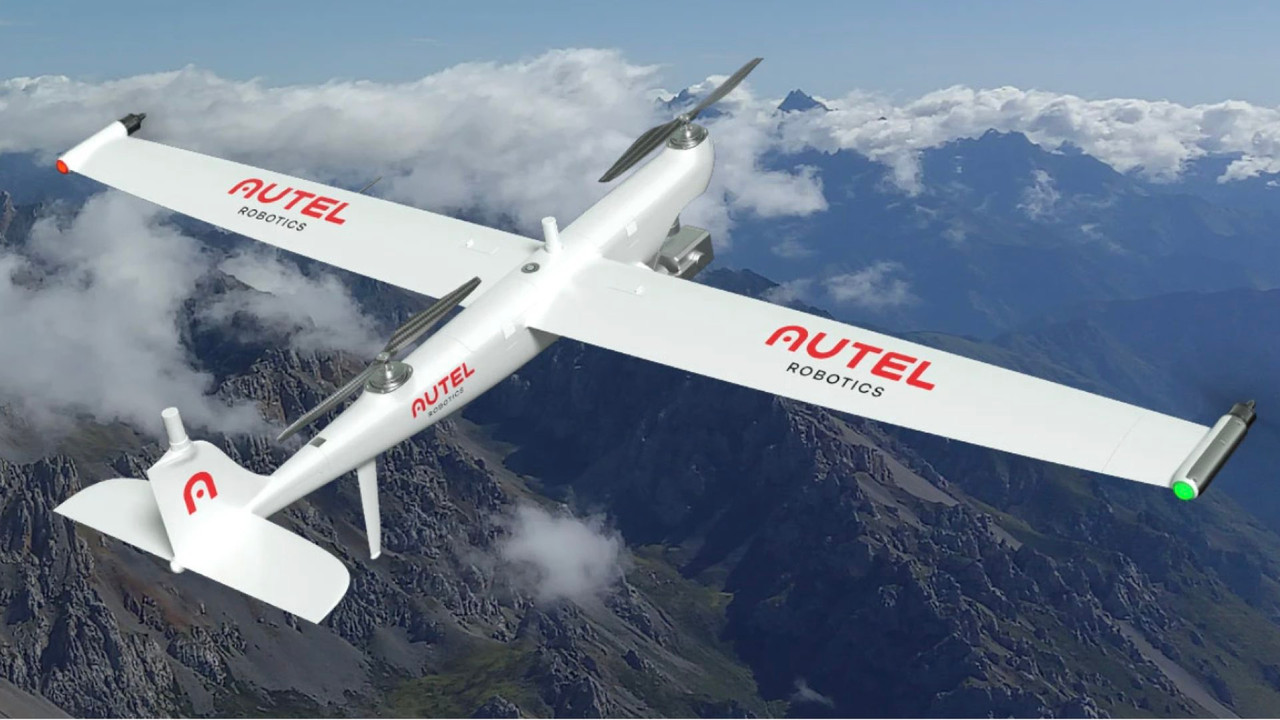}
    \caption{Example Fixed-Wing Drone - Autel Dragonfish}
    \label{fig:fixed_example}
    \end{figure}

    \item \textbf{Hybrid Drones}: Hybrid drones (Figure \ref{fig:hybrid_example}) combine the features of multi-rotor and fixed-wing drones \cite{10.1109/access.2021.3130900}. They can take off and land vertically like multi-rotor drones and then transition to fixed-wing flight for longer endurance and coverage \cite{10.1109/access.2021.3130900}. Hybrid drones are suitable for applications that require both close-range imaging and large-scale mapping, providing flexibility and versatility in agricultural operations.

    \begin{figure}[htbp]
    \centering
    \includegraphics[width=\columnwidth]{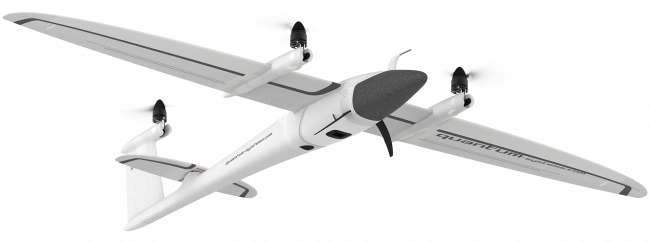}
    \caption{Example Hybrid Drone - Quantum Systems Trinity F90+ eVTOL Fixed-Wing UAV}
    \label{fig:hybrid_example}
    \end{figure}

    \item \textbf{Thermal Imaging Drones}: Thermal imaging drones (Figure \ref{fig:thermal_example}) are equipped with thermal cameras that capture infrared radiation emitted by objects \cite{10.1007/s41666-020-00080-6}. These drones are used in agriculture to monitor crop health, detect irrigation issues, and identify areas of heat stress or pest infestation \cite{10.1038/s41598-020-67898-3}. Thermal imaging drones can provide valuable insights into the temperature distribution and thermal patterns in agricultural fields, aiding precision agriculture practices.

    \begin{figure}[htbp]
    \centering
    \includegraphics[width=0.9\columnwidth]{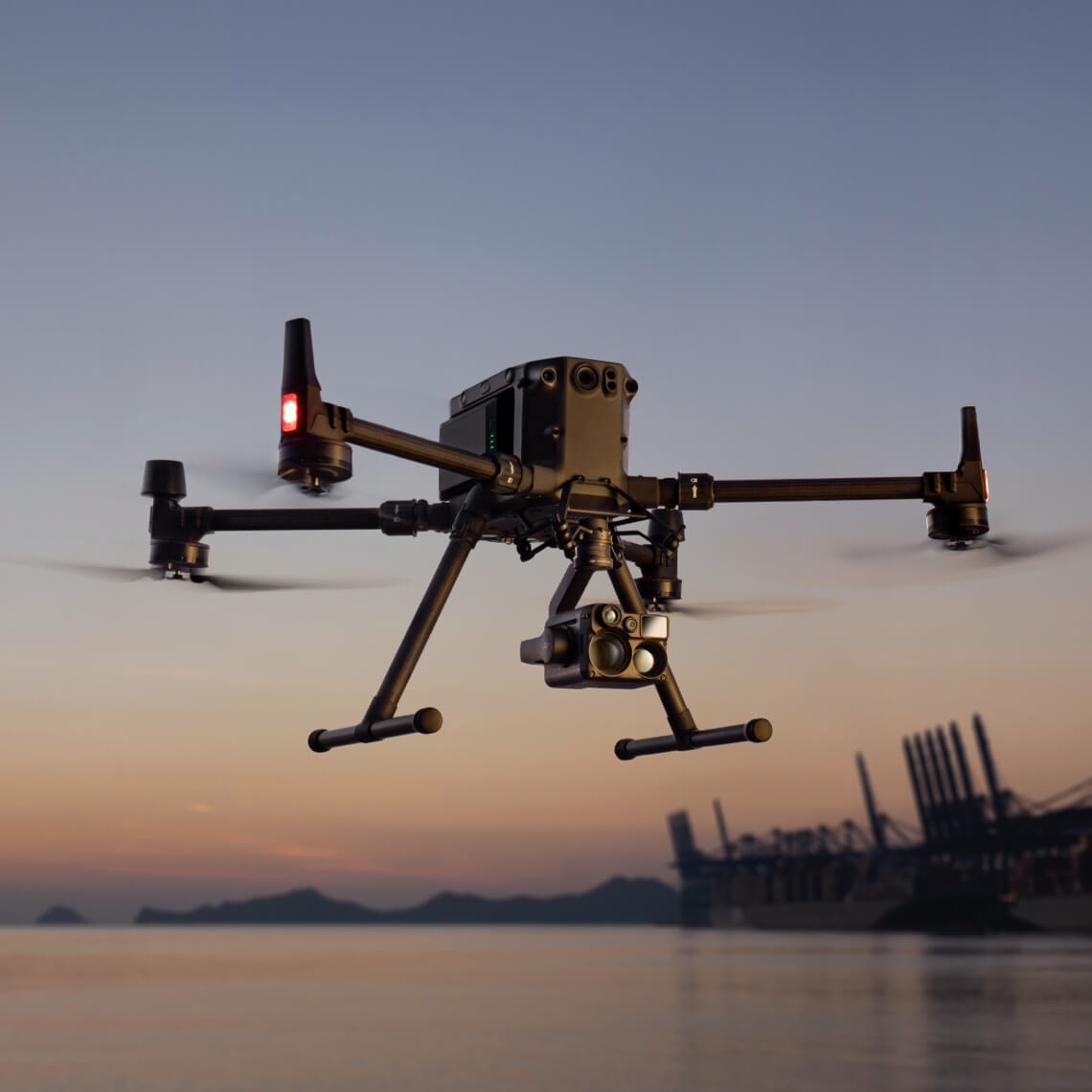}
    \caption{Example Hybrid Drone - DJI Zenmuse XT2}
    \label{fig:thermal_example}
    \end{figure}

    \item \textbf{Spraying Drones}: Spraying drones (Figure \ref{fig:spraying_example}), also known as agricultural drones or crop dusting drones, are specifically designed for the targeted application of pesticides, fertilizers, and other agricultural inputs \cite{10.1088/1757-899x/1259/1/012015}. These drones are equipped with spraying systems that can accurately and efficiently deliver chemicals to crops, reducing the need for manual labor and minimizing chemical wastage \cite{10.3390/app11052138}. Spraying drones offer precise and controlled applications, reducing environmental impact and optimizing resource utilization.

    \begin{figure}[htbp]
    \centering
    \includegraphics[width=\columnwidth]{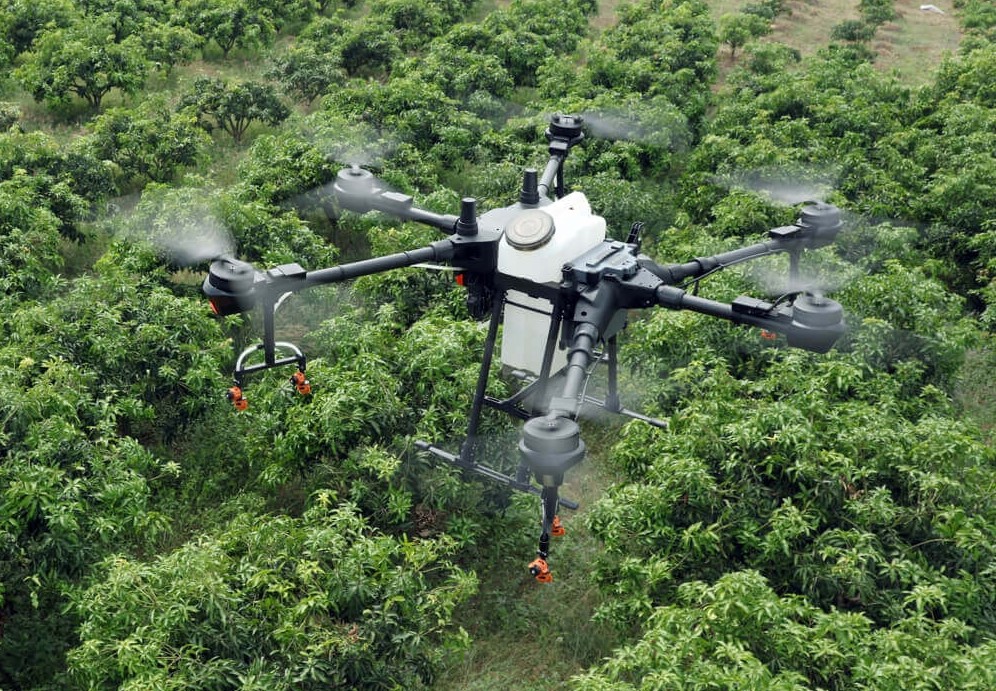}
    \caption{Example Spraying Drone - DJI Agras T16}
    \label{fig:spraying_example}
    \end{figure}

    \item \textbf{Surveillance Drones}: Surveillance drones (Figure \ref{fig:surveillance_example}) are used in agriculture for monitoring and security purposes \cite{10.1007/s41666-020-00080-6}. These drones are equipped with cameras and sensors that capture real-time video footage and imagery, allowing farmers to monitor their fields, livestock, and infrastructure remotely \cite{10.1109/access.2020.2982086}. Surveillance drones can help detect unauthorized activities, track animal movements, and identify potential threats or risks in agricultural operations.

    \begin{figure}[htbp]
    \centering
    \includegraphics[width=0.8\columnwidth]{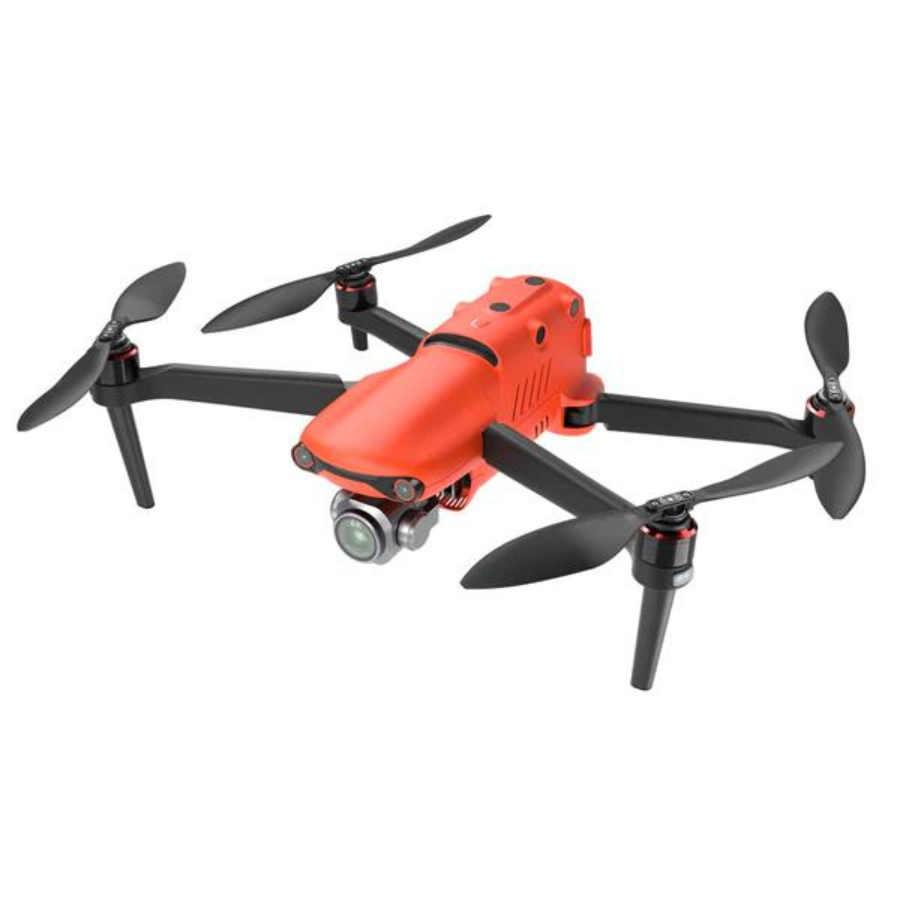}
    \caption{Example Surveillance Drone - Autel EVO II}
    \label{fig:surveillance_example}
    \end{figure}

    \item \textbf{Mapping and Surveying Drones}: Mapping and surveying drones (Figrue \ref{fig:mapping_example}) are used to create high-resolution maps and 3D models of agricultural fields \cite{10.3390/rs9010088}. These drones have advanced sensors, such as LiDAR (Light Detection and Ranging) or photogrammetry cameras, to capture detailed and accurate data \cite{10.3390/rs9010088}. Mapping and surveying drones are valuable tools for precision agriculture, enabling farmers to analyze topography, monitor soil conditions, and plan efficient land management strategies.

    \begin{figure}[htbp]
    \centering
    \includegraphics[width=0.8\columnwidth]{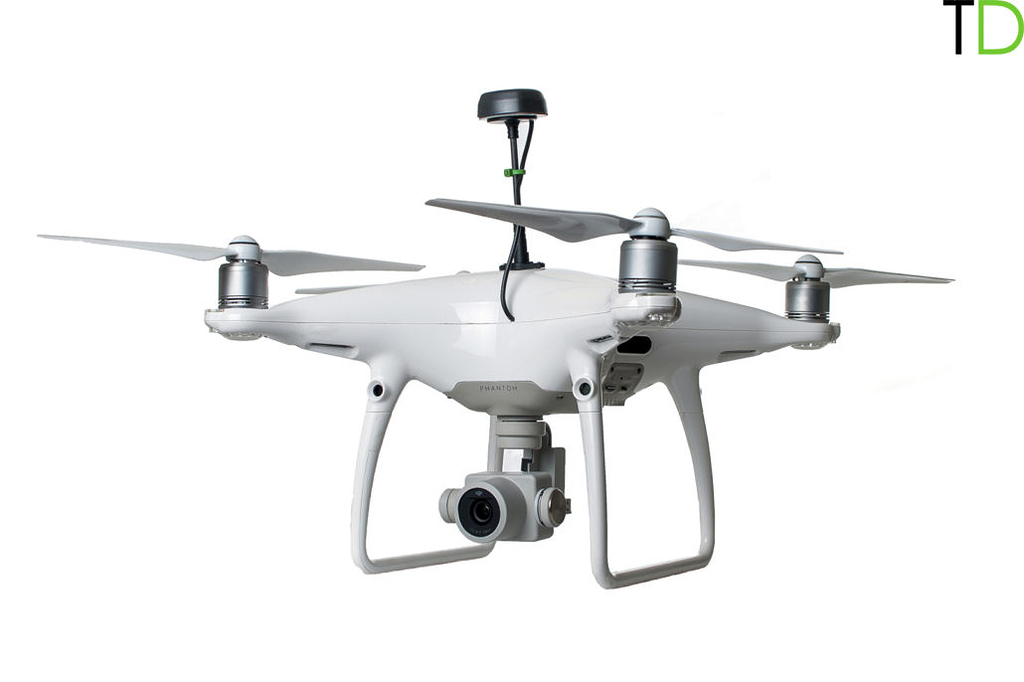}
    \caption{Example Mapping Drone - DJI Phantom 4 RTK}
    \label{fig:mapping_example}
    \end{figure}

    \item \textbf{Payload-Specific Drones}: Drones are designed for specific agricultural applications besides the above types. For example, there are drones equipped with hyperspectral sensors for detailed analysis of crop health and nutrient content \cite{10.3390/rs9010088}. There are also drones with specialized sensors for monitoring soil moisture levels, detecting weed infestations, or assessing plant growth parameters \cite{10.1111/sum.12771}. These payload-specific drones (Figure \ref{fig:mpaylod_example}) cater to specific data collection needs in agriculture.

    \begin{figure}[htbp]
    \centering
    \includegraphics[width=0.8\columnwidth]{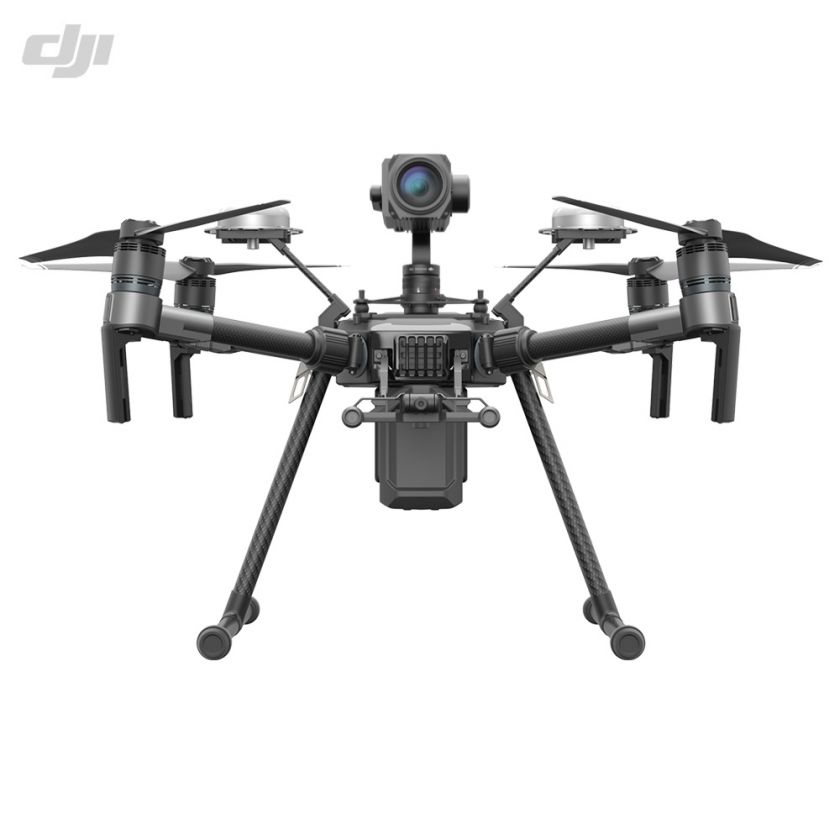}
    \caption{Example Payload-Specific Drone - DJI Matrice 210 RTK}
    \label{fig:mpaylod_example}
    \end{figure}
    
\end{itemize}

\section{Advantages of using Drones in Agriculture}

Using drones in agriculture offers several advantages contributing to improved efficiency, productivity, and sustainability in agricultural practices. The advantages of using drones in farming are: 

\begin{itemize}
    \item \textbf{Precision Agriculture}: Drones enable precision agriculture practices by providing high-resolution imagery and data collection capabilities \cite{10.1371/journal.pone.0141006}. They can capture detailed information about crop health, soil conditions, and pest infestations, allowing farmers to make informed decisions and apply targeted treatments \cite{10.1002/net.21818}. This precision approach helps optimize resource utilization, reduce input wastage, and increase crop yields \cite{10.1016/j.jairtraman.2020.101929}. 

    \item \textbf{Cost and Time Savings}: Drones can cover large areas of farmland quickly and efficiently, reducing the time and labor required for manual inspections and data collection \cite{10.1371/journal.pone.0141006}. They can perform tasks such as crop monitoring, mapping, and spraying in a fraction of the time it would take using traditional methods \cite{10.22438/jeb/43/1/mrn-1912}. This leads to cost savings by minimizing the need for manual labor and reducing the use of resources such as water, fertilizers, and pesticides \cite{10.3390/agronomy11091809}. 
    
    \item \textbf{Improved Data Collection and Analysis}: Drones equipped with various sensors, such as cameras, thermal imaging, and multispectral sensors, can collect a wide range of data about crops, soil, and environmental conditions \cite{10.5194/isprs-archives-xlii-2-789-2018}. This data can be used for detailed analysis and monitoring, enabling farmers to detect early signs of crop stress, nutrient deficiencies, or disease outbreaks \cite{10.1038/s41598-020-67898-3}. The data collected by drones can be processed using advanced analytics and machine learning algorithms to generate actionable insights for better decision-making \cite{10.30657/pea.2021.27.10}. 

    \item \textbf{Enhanced Crop Management}: Drones provide real-time and up-to-date information about crop health, allowing farmers to implement timely interventions and optimize crop management practices \cite{10.3390/drones5020041}. For example, drones can help identify areas of the field that require additional irrigation or fertilization, enabling precise application and reducing waste \cite{10.1007/978-981-16-4369-9_25}. They can also assist in monitoring crop growth, estimating yield potential, and predicting harvest times \cite{10.3390/agriculture13051075}. 

    \item \textbf{Accessibility and Flexibility}: Drones offer accessibility to areas that are difficult to reach or inaccessible by traditional means, such as steep slopes or dense vegetation \cite{10.5937/ekonomika1804091s}. They can fly at low altitudes and capture data from different angles and perspectives, providing a comprehensive view of the field \cite{10.1051/matecconf/202133502002}. Drones can be deployed quickly and easily, allowing farmers to respond rapidly to changing conditions or emergencies \cite{10.1051/e3sconf/202338101048}. 

    \item \textbf{Environmental Sustainability}: Using drones in farming can contribute to environmental sustainability by reducing the use of chemicals and minimizing the environmental impact of agricultural practices \cite{10.1016/j.jairtraman.2020.101929}. Drones enable targeted spraying of pesticides and fertilizers, reducing the amount of chemicals applied and minimizing their dispersion into the environment \cite{10.22438/jeb/43/1/mrn-1912}. This targeted approach helps protect beneficial insects, reduce water pollution, and promote ecological balance \cite{10.1038/s41598-020-67898-3}. 

    \item \textbf{Safety}: Drones eliminate or reduce the need for farmers to physically access hazardous or difficult-to-reach areas, such as tall crops, steep terrains, or areas with potential safety risks \cite{10.1002/net.21818}. This improves the safety of farmers and reduces the risk of accidents or injuries associated with manual labor \cite{10.3390/drones5020041}. 

\end{itemize}

\section{Conclusion}

Using drones in agriculture holds immense promise for revolutionizing farming practices and improving efficiency, productivity, and sustainability. The various types of drones available cater to specific agricultural needs, ranging from crop monitoring and assessment to precision spraying, mapping, and surveying. Drones provide real-time and high-resolution data collection, enabling farmers to make informed decisions regarding resource allocation and optimize crop management practices. They offer cost and time savings by reducing manual labor and minimizing the use of resources. The ability of drones to access inaccessible areas and provide comprehensive views of the fields enhances their usability and efficiency in large-scale farming operations.
Furthermore, drones contribute to environmental sustainability by enabling targeted spraying, reducing chemical wastage, and minimizing the environmental impact of agricultural practices. The safety aspect of using drones must be considered, as they eliminate or reduce the need for farmers to access hazardous areas physically. Despite challenges such as regulations and limited flight time, advancements in drone technology are continually addressing these limitations. Overall, the advantages of using drones in agriculture are significant, and their integration into farming practices has the potential to transform the industry, leading to optimized resource utilization, improved crop yields, and sustainable agricultural practices.

\end{document}